\documentclass{svproc}
\usepackage{graphicx}
\usepackage{booktabs}

\usepackage[utf8]{inputenc} %
\usepackage[T1]{fontenc}    %
\usepackage{hyperref}       %
\usepackage{url}            %

\usepackage{amsfonts}       %
\usepackage{nicefrac}       %
\usepackage{microtype}      %
\usepackage{multirow}
\usepackage{xcolor}
\usepackage{amsmath}
\usepackage{cellspace}
\usepackage{pifont}
\usepackage{booktabs}
\usepackage{mwe}
\usepackage{tabularx}
\usepackage[inline]{enumitem}
\usepackage{array}
\usepackage{makecell}

\newcolumntype{H}{>{\setbox0=\hbox\bgroup}c<{\egroup}@{}}
\newcommand{\para}[1]{\noindent\textbf{#1}}

\newcommand\blfootnote[1]{%
  \begingroup
  \renewcommand\thefootnote{}\footnote{#1}%
  \addtocounter{footnote}{-1}%
  \endgroup
}

\usepackage{url}

\begin{document}
\mainmatter              %
\title{On the Impact of Communities on Semi-supervised Classification Using Graph Neural Networks}
\titlerunning{Communities in Semi-supervised Classification with GNNs}
\author{Hussain Hussain\inst{1} \and
Tomislav Duricic\inst{1,2}\and
Elisabeth Lex\inst{1,2} \and
Roman Kern\inst{1,2} \and
Denis Helic\inst{2}}
\authorrunning{H. Hussain et al.}
\institute{
Know-Center GmbH
\and
Graz University of Technology 
\email{\{hussain,tduricic,elisabeth.lex,rkern,dhelic\}@tugraz.at}
}
\maketitle 
\begin{abstract}
Graph Neural Networks (GNNs) are effective in many applications. Still, there is a limited understanding of the effect of common graph structures on the learning process of GNNs. In this work, we systematically study the impact of community structure on the performance of GNNs in semi-supervised node classification on graphs. Following an ablation study on six datasets, we measure the performance of GNNs on the original graphs, and the change in performance in the presence and the absence of community structure. Our results suggest that communities typically have a major impact on the learning process and classification performance. For example, in cases where the majority of nodes from one community share a single classification label, breaking up community structure results in a significant performance drop. On the other hand, for cases where labels show low correlation with communities, we find that the graph structure is rather irrelevant to the learning process, and a feature-only baseline becomes hard to beat. With our work, we provide deeper insights in the abilities and limitations of GNNs, including a set of general guidelines for model selection based on the graph structure.\blfootnote{This is a preprint of the following chapter: Hussain H., Duricic T., Lex E., Kern R., Helic D., On the Impact of Communities on Semi-supervised Classification Using Graph Neural Networks, published in Complex Networks \& Their Applications IX, edited by Rosa M. Benito, Chantal Cherifi, Hocine Cherifi, Esteban Moro, Luis Mateus Rocha, Marta Sales-Pardo, 2020, Springer, Cham reproduced with permission of Springer, Cham. The final authenticated version is available online at: \url{http://dx.doi.org/10.1007/978-3-030-65351-4_2}.}

 \end{abstract}
%This is a preprint of the following chapter: Hussain H., Duricic T., Lex E., Kern R., Helic D., On the Impact of Communities on Semi-supervised Classification Using Graph Neural Networks, published in Complex Networks & Their Applications IX, edited by Rosa M. BenitoChantal CherifiHocine CherifiEsteban MoroLuis Mateus RochaMarta Sales-Pardo, 2020, Springer, Cham reproduced with permission of Springer, Cham. The final authenticated version is available online at: \url{http://dx.doi.org/10.1007/978-3-030-65351-4_2}.

\section{Introduction}

Many real-world systems are modeled as complex networks, which are defined as graphs with complex structural features that cannot be observed in random graphs~\cite{kim2008complex}.
The existence of such features governs different processes and interactions between nodes in the graph.
In particular, community structures are often found in empirical real-world complex networks. These structures have a major impact on information propagation across graphs~\cite{barabasi2013network} as they provide barriers for propagation~\cite{hasani2018consensus}.
The process of information propagation forms the basis for many studied applications on graphs.
Among these applications, graph-based semi-supervised learning~\cite{gcn,zhu2003semi} is widely studied and particularly of interest.
Graph-based semi-supervised learning aims to exploit graph structure in order to learn with very few labeled examples~\cite{Li2018Deeper}.
In recent years, state-of-the-art methods for this task have predominantly been graph neural networks (GNNs)~\cite{shchur2018pitfalls}. 

\para{Problem.} Despite the outstanding results of GNNs on this and similar tasks, there is still a limited understanding of their abilities and constraints, which hinders further progress in the field~\cite{Loukas2020What,Xu2019How}. To the best of our knowledge, there is a lack of research on how common graph structures, such as communities, impact the learning process of GNNs.

\para{The present work.}
We are particularly interested in the influence of community structure on the performance of GNNs in semi-supervised node classification.
From a practical perspective, we set out to provide a set of guidelines on the applicability of GNNs based on the relationships between communities and target labels.
To that end, we design an evaluation strategy based on an ablation study on multiple graph datasets with varying characteristics in order to study the behaviour of GNNs.
Using this evaluation strategy we compare the performance of GNNs on six public graph datasets to a simple feature-based baseline (i.e., logistic regression) which ignores the graph structure.

To gain a deeper understanding on the role of communities, we compare the change in performance of GNNs after (i) eliminating community structures while keeping the degree distribution, (ii) keeping the community structure while using a binomial degree distribution, and (iii) eliminating both.
Finally, we link the evaluation results on the ablation models to the relationship between communities and labels in each dataset.
To achieve this, we compute the uncertainty coefficient~\cite{press1992numerical} of labels with respect to communities, before and after applying community perturbations.

\para{Findings.}
Our results show that GNNs are able to successfully exploit the graph structure and outperform the feature-based baseline only when communities correlate with the node labels, which is known in the literature as the cluster assumption~\cite{chapelle2009semi}.
If this assumption fails, GNNs propagate noisy features across the graph and are unable to outperform the feature-based baseline.

\para{Contributions.}
With our work, we highlight the limitations that community structures can impose on GNNs.
We additionally show that the proposed uncertainty coefficient measure helps predicting the applicability of GNNs on semi-supervised learning.
We argue that this measure can set a guideline to decide whether to use GNNs on a certain semi-supervised learning task, given the relationship between communities and labels.

\section{Background}
\para{Graph neural networks.}
Let $G=(V,E)$ be a graph with a set of nodes $V$ and a set of edges $E$.
Let each node $u \in V$ have a feature vector $x_u \in I\!\!R^d$, where $d$ is the feature vector dimension; and a label $y_u \in \mathcal{L}$, where $\mathcal{L}$ is the set of labels.

GNNs are multi-layer machine learning models, which operate on graphs.
They follow a message passing and aggregation scheme where nodes aggregate the messages that are received from their neighbors and update their representation on this basis.
In a GNN with $K$ hidden layers (with the input layer denoted as layer $0$), each node $u$ has a vector representation $h_u^{(k)}$ at a certain layer $k \leq K$ of dimension $d_k$.
The transformation from a layer $k$ to the next layer $k+1$ is performed by updating the representation of each node $u$ as follows:
\begin{equation}
    \label{eq:gnn}
    \begin{split}
    a_u^{(k)} & := \text{AGGREGATE}_{v \in \mathcal{N}(u)}(h_v^{(k)}),\\
    h_u^{(k+1)} & := \text{COMBINE}(h_u^{(k)},a_u^{(k)}),
    \end{split}
\end{equation}
where $\mathcal{N}(u)$ is the set of neighbors of $u$.
The AGGREGATE function takes an unordered set of vectors of dimension $d_k$ as an input, and returns a single vector of the same dimension $d_k$, e.g., element-wise mean or max.
The COMBINE function combines the representation of $u$ in layer $k$ with the aggregated representation of its neighbors, e.g.,  a concatenation followed by ReLU of a linear transformation $\text{COMBINE}(h,a) = \text{ReLU}(W.[h,a])$.
We set the representation of $u$ in the input layer to the input features: $h_u^{(0)} := x_u$.
In classification problems, the dimension of the last layer $d_K$ equals the number of labels in the graph $|\mathcal{L}|$.

\para{Semi-supervised learning on graphs.}
Semi-supervised learning aims to exploit unlabeled data in order to generate predictions given few labeled data.
In graphs, this means exploiting the unlabeled nodes as well as the network structure to improve the predictions.
Many semi-supervised classification methods on graphs assume that connected nodes are more likely to share their label~\cite{Li2018Deeper}, which is usually referred to as the \textit{cluster assumption}~\cite{chapelle2009semi}.  
Based on this assumption, approaches to solve this task usually aim to propagate node information along the edges.
Earlier related approaches~\cite{sen2008collective,zhu2003semi} focused on propagating label information from labeled nodes to their neighbors.
In many applications, however, graph nodes can also be associated with feature vectors, which can be utilized by GNNs.
GNNs achieved a significant improvement over the state of the art since they can effectively harness the unlabeled data, i.e., graph structure and node features.

\para{Cluster assumption.} GNNs operate by propagating node feature vectors along the edges, hence exploiting both the graph structure and feature vectors. %
The GNN update rule in Equation~\ref{eq:gnn} can be seen as a form of \textit{(Laplacian) feature smoothing}~\cite{Li2018Deeper} as it combines the feature vector of a node with the feature vectors of its neighbors.
Feature smoothing results in neighboring nodes having similar vector representations.
Therefore, with the cluster assumption in mind, feature smoothing potentially causes nodes with similar labels to also obtain similar vector representations.
However, when the cluster assumption does not hold, i.e., connected nodes are less likely to share their label, the propagation in Equation~\ref{eq:gnn} can cause nodes with different labels to have similar vector representations.
It is widely accepted that classifiers achieve better accuracy when similar vector representations tend to have similar labels.

\para{Communities and the cluster assumption.}
Communities are densely connected subgraphs, and they are common in empirical graphs including social, citation or web graphs.
The existence of communities directly affects information propagation in graphs~\cite{cherifi2019community}.
As communities are densely connected, the feature smoothing performed by the update rule in Equation~\ref{eq:gnn} tends to make nodes within the same community have similar vector representations.
This dense connectivity also causes the cluster assumption to generalize to the community level, which means that nodes within the same community tend to share the same label.
As a result, when the cluster assumption holds, GNNs cause nodes with the same label to have similar vector representations simplifying the classification task on the resulting vector representations. Li et al.~\cite{Li2018Deeper} hypothesize that this alignment of communities and node labels may be the main reason why GNNs achieve state-of-the-art performance on the classification task. In this paper we aim to experimentally test this hypothesis on six datasets from different domains. 

In the other case, which is typically ignored in literature, the cluster assumption does not hold, and a community could possibly have a variety of labels.
The feature propagation between nodes of the same community would therefore result in feature smoothing for nodes with different labels.
This eventually makes the classification task harder since representation similarity does not imply label similarity in this case.

In summary, in this paper we set out to quantify the label-community correlation and how this correlation is related to the performance of GNNs on semi-supervised classification task on graphs.

\section{Methods and Experiments}

We start by quantifying how much information node's community reveals about its label.
For a labeled graph, let $L$ be a random variable taking values in the set of labels $\mathcal{L}$, i.e., $L(u)$ is the label of node $u \in V$.
Assuming the graph is partitioned into a set of disjoint communities $\mathcal{C}$, we define another random variable $C$ taking values in $\mathcal{C}$, i.e., $C(u)$ is the community of node $u \in V$.

To measure how much the (fraction of) uncertainty about $L$ is reduced knowing $C$, we use the uncertainty coefficient~\cite{press1992numerical} of $L$ given $C$.
This coefficient can be written as $U(L|C)=\frac{I(L;C)}{H(L)} \in [0,1]$, where $H(L)$ is the entropy of  $L$, and $I(L;C)$ is the mutual information between $L$ and $C$.

When the uncertainty coefficient equals $1$, all nodes within each community share the same label, and thus knowing the node's community means also that we know the node's label. On the other hand, when the uncertainty coefficient is $0$, the label distribution is identical in all communities, so knowing the community of a node does not contribute to knowing its label. In general, the higher the eliminated uncertainty about the labels when knowing communities is (i.e., the closer $U(L|C)$ is to $1$), the more likely it is that the cluster assumption holds, and thus GNNs can exploit the graph structure, and vice versa.%

\subsection{Ablation study}
\label{sec:ablation}
After establishing the intuitions behind the role of communities, we aim to show their impact experimentally.
To achieve this, we evaluate five popular state-of-the-art GNN models on six empirical datasets. Subsequently, we re-evaluate these GNN models on the same datasets after applying ablation to their structures.

In particular, we evaluate the accuracy of GNNs on the original datasets and compare this performance to the ones on the following ablation models:
\begin{itemize}
    \item \textbf{SBM networks.} Here we aim to rebuild the graph while preserving the communities. Thus, we firstly perform community detection with widely used Louvain method~\cite{louvain}, which maximizes the modularity score. Secondly, we build a stochastic block matrix encoding the original density of edges within and between the detected communities. Finally, we use this matrix to construct a graph with the stochastic block model (SBM)~\cite{holland1983stochastic}. This graph preserves a node's community, features and label but results in a binomial degree distribution~\cite{karrer2011stochastic}.%
    \item \textbf{CM networks.} In this ablation model, we apply graph rewiring using the so called configuration model (CM)~\cite{confmodel}.
    With this rewiring, each node keeps its degree, but its neighbors can become any of the nodes in the graph.
    This effectively destroys the community structure, while keeping the node's degree, features and label.
    \item \textbf{Random networks.} We use Erd{\H{o}}s-R{\'e}nyi graphs~\cite{erdos1960evolution} to eliminate both communities and degree distribution.
    The resulting graph has no community structure and features a binomial degree distribution.
This ablation model can only spread noisy feature information across the graph.
\end{itemize}

Last but not least, for each of the original datasets and each of the ablation models we compute the community-label correlation by means of the uncertainty coefficient.
To that end, we use the joint distribution of labels $L$ and communities $C$ (extracted by the Louvain method) to compute this coefficient for each graph.
In order to highlight the correlation between this coefficient and the applicability of GNNs, we show the computed coefficients on the given datasets along with their performance.

\subsection{Experiments}
In our experiments~\footnote{The implementation and technical details can be found on \url{https://github.com/sqrhussain/structure-in-gnn}}, we study five GNN architectures which are widely used for semi-supervised classification on graphs:
\begin{enumerate*}[label=(\alph*)]
    \item Graph Convolutional Networks (GCN)~\cite{gcn},
    \item Graph Sample and Aggregate (SAGE)~\cite{graphsage},
    \item Graph Attention Networks (GAT)~\cite{gat},
    \item Simple Graph Convolutions (SGC)~\cite{sgc}, and
    \item Approximate Personalized Propagation of Neural Predictions (APPNP)~\cite{Klicpera2019predict}%
\end{enumerate*}.
We additionally compare these approaches to a simple feature-only baseline, i.e., logistic regression model, which ignores the graph structure and only uses the node features.
The comparison to this baseline can indicate whether a GNN model is actually useful for the task on the respective datasets.

\para{Datasets.}
\label{sec:datasets}
To provide a better understanding of the roles of the studied structures, we aim for a diverse selection of datasets with respect to
\begin{enumerate*}[label=(\alph*)]
    \item domain, e.g., citation, social and web graphs,
    \item structure, e.g., directed acyclic vs. cyclic,
    \item and correlations between communities and labels, i.e., whether nodes of the same community tend to share the same label. 
\end{enumerate*}
Having this in mind, we use the datasets summarized in Table~\ref{tab:datasets}.

\begin{table}[b!]
\scriptsize
\caption{Dataset statistics after preprocessing (similar to Shchur et al.~\cite{shchur2018pitfalls}). The label rate represents the fraction of nodes in the \textit{training} set. The edge density is the number of existing undirected edges divided by the maximum possible number of undirected edges (ignoring self-loops). For Twitter dataset, we apply cleaning to the feature vectors of nodes same as in~\cite{Tiao2019}, i.e., removing graph-based and some textual features. }
\label{tab:datasets}
\centering
\begin{tabular}{lrrrrrr}
\toprule
Dataset & Labels & Features & Nodes & Edges & Edge density & Label rate \\
\midrule
CORA-ML~\cite{sen2008collective} & 7 & 1,433 & 2,485 & 5,209 & 0.0017 & 0.0563 \\
CiteSeer~\cite{sen2008collective} & 6 & 3,703 & 2,110 & 3,705 & 0.0017 & 0.0569 \\
PubMed~\cite{pubmed} & 3 & 500 & 19,717 & 44,335 & 0.0002 & 0.0030 \\
CORA-Full~\cite{corafull} & 67 & 8,710 & 18,703 & 64,259 & 0.0004 & 0.0716 \\
Twitter~\cite{twitter} & 2 & 215 & 2,134 & 7,040 & 0.0031 & 0.0187 \\
WebKB~\cite{webkb} & 5 & 1,703 & 859 & 1,516 & 0.0041 & 0.0582 \\
\bottomrule
\end{tabular}
\end{table}

The label of a node in a \textit{citation graph} (i.e., \textit{CORA-ML, CiteSeer, PubMed} and \textit{CORA-Full}) represents the topic of the paper. Citations are expected to be denser among papers with similar topics than they would be between papers of different topics. For example, a publication about natural language processing would be more likely to cite other natural language processing papers than human-computer interaction papers. Therefore, one could intuitively expect that papers within the same graph community tend to share the same label.

\textit{Twitter} graph consists of users where the edges represent retweets, and node labels indicate whether a user is hateful or not.
Therefore, one could not easily assume the presence or the absence of a correlation between communities and labels (hateful or normal). In other words, we do not know whether hateful users typically form communities as this highly depends on the discussion topics.

For \textit{WebKB} dataset, nodes are web pages, edges are links, and labels indicate the type of the web page, i.e., course, faculty, project, staff or student.
In this case one cannot intuitively assume that nodes within a graph community are expected to share a label.
For example, a web page of a staff member could be more likely to link to projects on which this staff member is working than to other staff members' web pages.
Based on these intuitions, we consider that these graphs are sufficiently diverse concerning community impact on label prediction.%

\para{Evaluation setup}.
While the original graphs are directed, we treat them as undirected by ignoring the edge direction (which is in the line with previous research~\cite{graphsage,gcn,shchur2018pitfalls}).
All of these graphs are preprocessed in a similar manner as Shchur et al.~\cite{shchur2018pitfalls}, i.e., removing nodes with rare labels and selecting the largest weakly connected component except in WebKB where we take four connected components representing 4 universities.
Following the train/validation/test split strategy as in~\cite{shchur2018pitfalls}, each random split consists of $50$ labeled examples per class ($20$ for training and $30$ for validation) and the rest are considered test examples.
This applies to all of our datasets except WebKB where we use $10$ training examples and $15$ validation examples per class due to the fewer number of nodes.
To evaluate the GNN models on the original graphs, we follow the evaluation setup as conducted in~\cite{shchur2018pitfalls} by having $100$ random splits and $20$ random model initializations for each split.
The same process is carried out to evaluate the feature-only baseline (logistic regression) model.
The evaluation is slightly different for the ablation studies (SBM, CM and random graphs) since they include an additional level of randomization.
For a given dataset, we generate SBM, CM and random graphs with $10$ different random seeds, which results in $10$ SBM graphs, $10$ CM graphs and $10$ random graphs.
The evaluation on each of these generated graphs is carried out through $50$ different random splits and $10$ different random initializations.
As a result, the reported accuracy for a GNN architecture on the original graph is presented for $2,000$ trainings of the GNN.
Meanwhile, the reported accuracy of a GNN architecture on one of the ablation models (SBM, CM or random graphs) is presented for training the GNN $5,000$ times, i.e., $10$ randomly generated graphs $\times$ $50$ random splits $\times$ $10$ random initializations.

\para{Uncertainty coefficient calculation.}
To calculate $U(L|C)$ for a set of nodes, the labels of these nodes must be available.
Therefore, for each dataset, we compute $U(L|C)$ using the labeled nodes from the training and validation sets.%

\subsection{Results}

\begin{figure}[t!]
    \centering
    \includegraphics[width=0.95\textwidth]{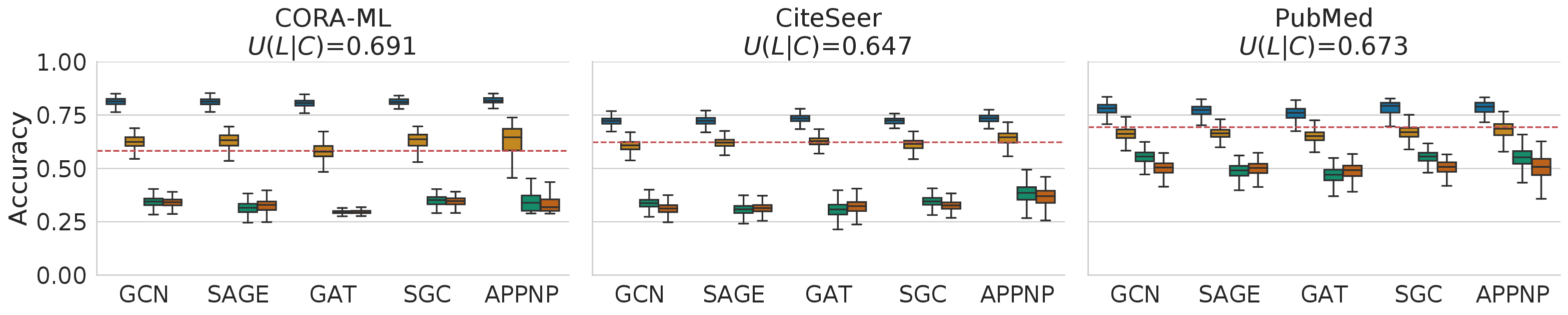}
    \includegraphics[width=0.95\textwidth]{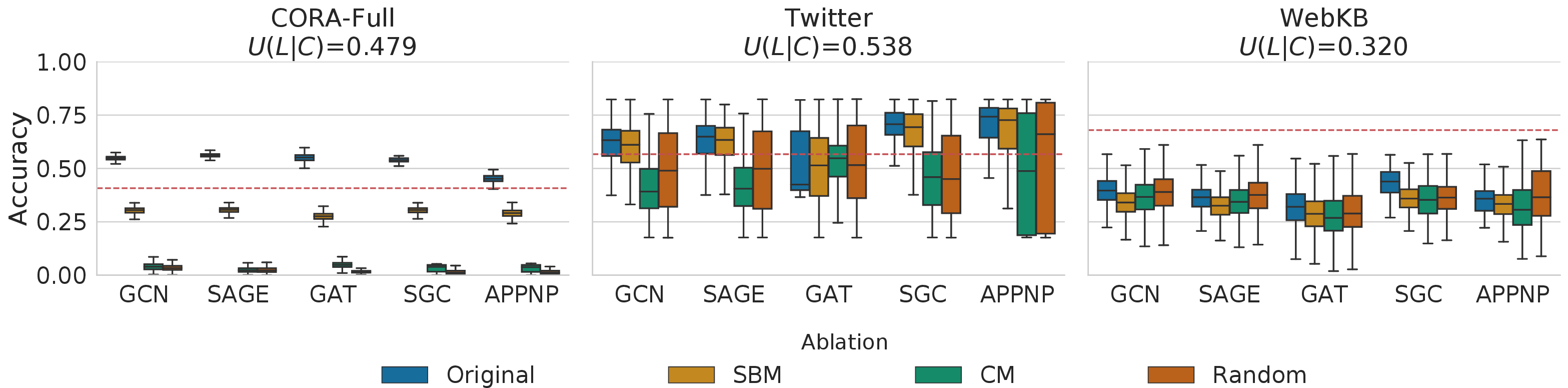}
    
    \caption{The accuracy of GNNs on the original graph and the ablation models of each dataset.
    The red dashed line represents the median accuracy of the feature-only baseline.
    The performance on the original graphs is generally higher than that of the baseline except for WebKB.
    On citation graphs, the baseline is clearly outperformed on the original graphs, and eliminating communities (CM graphs) results in a much higher accuracy drop than eliminating the degree sequence (SBM graphs).
    This is not the case for the other two datasets where the baseline is not always outperformed on the original graphs, and the drop in performance on SBM networks is substantial. This highlights the positive impact of communities on citation datasets and its negative impact on WebKB. %
    The uncertainty coefficient is the highest for the three datasets in the top row and the lowest for WebKB providing a potential explanation for the low GNN performance on this dataset.
    }
    \label{fig:results}
\end{figure}

We summarize the evaluation results of the GNN models with respect to accuracy for the original graph and its corresponding ablation models in Figure~\ref{fig:results}.

\para{Could GNNs outperform the simple baseline?} 
Comparing the performance of GNNs on the original graphs with the feature-only baseline, all GNN models clearly outperform the baseline on the citation datasets. For the Twitter dataset however, GAT could not outperform the baseline, while GCN and GraphSAGE outperformed it only by a small margin. 
This is more prominent for WebKB where none of the GNNs outperforms the baseline on the original graph.
This suggests that for the two latter datasets, the graph structure is either irrelevant to the learning process or is even hindering it. To test the statistical significance for each approach on each dataset against the corresponding baseline, we compute the non-parametric \textit{Mann-Whitney U test} for unpaired data with the significance level $\alpha=0.01$ (Bonferroni corrected).
The significance test shows that the performance of GNNs on the original graphs is significantly different than the performance of the feature-only baseline, i.e., GNNs significantly outperform the baseline on all datasets except on WebKB where the baseline significantly outperforms the GNNs.
Mapping our results back to the cluster assumption and the uncertainty coefficient, we notice that when $U(L|C)$ is high, i.e., for CORA-ML (.691), CiteSeer (.647) and PubMed (.673), GNNs are able to consistently outperform the baseline. Meanwhile, when $U(L|C)$ is low, i.e., for WebKB (.320), GNNs are not useful for the task.%

\para{Ablation results.} On citation datasets, the accuracy drop on the SBM graphs is smaller than for the other two ablation models (CM and random graphs).
As SBM graphs preserve the community structure, this observation shows a noticeable impact of communities on node classification in citation datasets. However, this behavior is not always demonstrated on the Twitter dataset.
On contrary, on the Twitter dataset, we observe an overlap in the performances on the original graph and the ablation models. This performance overlap is even more prominent on WebKB.
The SBM graphs generally gain the lowest accuracy on the WebKB dataset, showing a negative effect of preserving communities in this dataset.
These observations can again indicate that communities are boosting the prediction for the citation datasets (where $U(L|C)$ is high) while hindering it for the WebKB dataset (where $U(L|C)$ is low).%

\subsection{Discussion}
\label{sec:uncertainty-result}
In our experiments, we observe that the uncertainty coefficient is high for the citation datasets, relatively low for Twitter, and much lower for WebKB.
This correlates with the classification performance on citation datasets that show (a) a consistent performance on the original graphs and (b) a better accuracy on SBM graphs comparing to the other ablations.
On contrary, in the Twitter dataset, where GNNs only outperform the baseline by a small margin, the behavior is reflected in the coefficient value that is smaller than in three out of four citation datasets.
For WebKB, the coefficient is particularly low following the observations that GNNs are unable to beat the simple baseline on the original network.

Our observation suggest that the uncertainty coefficient can indicate whether GNNs are applicable depending on the relationship between the communities and labels.
To shed more light on the correlation between the uncertainty coefficient and the classification performance of GNNs, we now study the change of both GNN accuracy and the measured uncertainty coefficient on the given datasets after applying additional community perturbations.
Particularly, we start with the SBM networks for each dataset and we perform the following perturbations.
We randomly select a fraction of nodes and assign them to different communities by simply swapping the nodes position in the network.
Then we gradually increase the fraction of the selected nodes to obtain a spectrum of the uncertainty coefficient.
Finally, we compute $U(L|C)$ and the accuracy of the GCN model on each of the obtained graphs and show the correlation between the two measures.
We choose the SBM networks for this experiment to guarantee that the node swapping only changes nodes' communities and not their importance.
We expect that these perturbations increasingly reduce $U(L|C)$ when the cluster assumption holds.

\begin{figure}[t!]
    \centering
    \includegraphics[width=\linewidth]{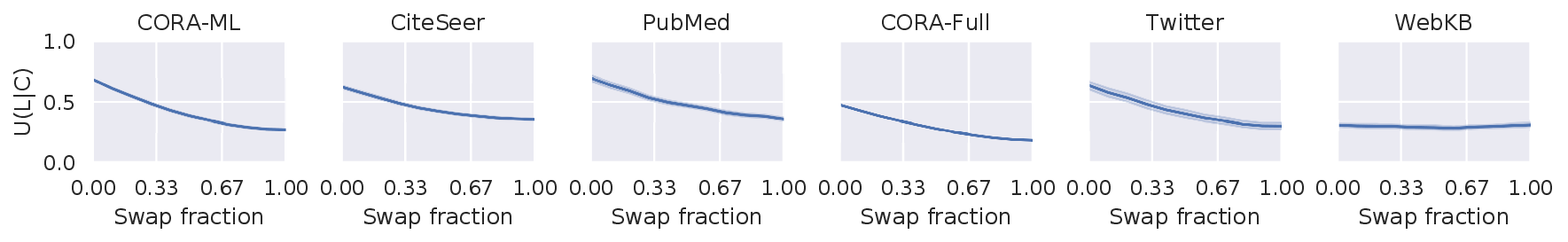}
    \includegraphics[width=\linewidth]{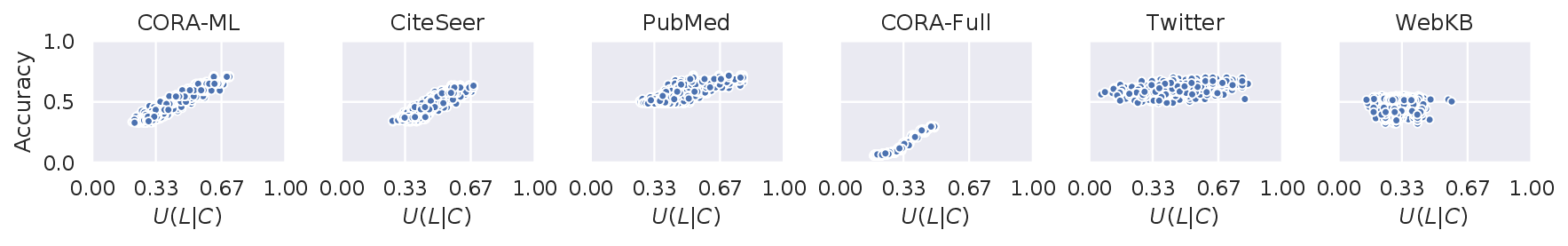}
    \caption{The figure in the top row shows the calculated $U(L|C)$ for each dataset. The swap fraction represents the fraction of nodes which changed their community. We see a decline in the coefficient with increasing swapping fraction, and then a convergence. This line is already converging in WebKB as the measure is already low in the original network. The figure in the bottom shows the test accuracy with changing $U(L|C)$. We see a positive correlation between the uncertainty coefficient and the accuracy, which is weak for Twitter, and non-existent for WebKB supporting our observations from above.}
    \label{fig:swap}
\end{figure}

We show that $U(L|C)$ decreases and then converges for the first 5 datasets datasets (cf. Figure~\ref{fig:swap} [top]), supporting that the community perturbations decrease the correlation between communities and labels.
In these cases, the GNN accuracy has a positive correlation with $U(L|C)$ (cf. Figure~\ref{fig:swap} [bottom]).
However, when community perturbations do not reduce the correlation between communities and labels, $U(L|C)$ is already at a convergence level.
That is the case where GNNs were not able to outperform the feature-based baseline, i.e., WebKB.
By reading the bottom row of Figure~\ref{fig:swap}, we see that when $U(L|C)$ is below $0.3$, the accuracy becomes unacceptably low. On the contrary, if $U(L|C)$ is above $0.7$, the accuracy is high enough for the respective dataset.

\para{Guideline for application of GNNs.} %
To verify whether a GNN model is applicable on a certain real-world dataset, we suggest the following two-step guideline based on the previous observations.
\begin{itemize}
    \item The first step is to perform community detection on the dataset, and inspect the uncertainty coefficient for it.
If the coefficient is particularly low, e.g., below $0.3$, there is a high confidence that GNNs will not work.
If the coefficient is particularly high, e.g., above $0.7$, it is likely that GNNs can exploit the network structure and be helpful for the prediction task.
Otherwise, if the value is inconclusive, e.g., around $0.5$, we advise to perform the second step.
    \item The second step involves gradual community perturbations and inspection of the respective uncertainty coefficient.
If the value of the coefficient decreases with more perturbations, this supports that the cluster assumption holds in the original graph, and GNNs are applicable.
Otherwise, if it does not decrease, the cluster assumption most likely does not hold in the first place, and feature-based methods are more advisable than GNNs.
\end{itemize}

\section{Conclusion}

In this work, we analyzed the impact of community structures in graphs on the performance of GNNs in semi-supervised node classification, with the goal of uncovering limitations imposed by such structures.
By conducting ablation studies on the given graphs, we showed that GNNs outperform a given baseline on this task in case the cluster assumption holds. Otherwise, GNNs cannot effectively exploit the graph structure for label prediction.
Additionally, we show an analysis on the relation between labels and graph communities. With our analysis, we suggest that when community information does not contribute to label prediction, it is not advisable to use such GNNs.
In particular, we show that the uncertainty coefficient of node labels knowing their communities can indicate whether the cluster assumption holds.
We further formalize a guideline to select where to apply GNNs based on community-label correlation.
Our work serves as a contributing factor to intrinsic validation of the applicability of GNN models.
Future work can also investigate the effect of other graph structural properties such as edge directionality, degree distribution and graph assortativity on the GNN performance.

\bibliographystyle{splncs03}
\bibliography{references}

\begin{thebibliography}{10}
\providecommand{\url}[1]{\texttt{#1}}
\providecommand{\urlprefix}{URL }

\bibitem{barabasi2013network}
Barab{\'a}si, A.L.: Network science. Philosophical Transactions of the Royal
  Society A: Mathematical, Physical and Engineering Sciences  371(1987),
  20120375 (2013)

\bibitem{louvain}
Blondel, V.D., Guillaume, J.L., Lambiotte, R., Lefebvre, E.: Fast unfolding of
  communities in large networks. Journal of statistical mechanics: theory and
  experiment  2008(10),  P10008 (2008)

\bibitem{corafull}
Bojchevski, A., G{\"u}nnemann, S.: Deep gaussian embedding of graphs:
  Unsupervised inductive learning via ranking. In: International Conference on
  Learning Representations. pp. 1--13 (2018)

\bibitem{chapelle2009semi}
Chapelle, O., Scholkopf, B., Zien, A.: Semi-supervised learning (chapelle, o.
  et al., eds.; 2006)[book reviews]. IEEE Transactions on Neural Networks
  20(3),  542--542 (2009)

\bibitem{cherifi2019community}
Cherifi, H., Palla, G., Szymanski, B.K., Lu, X.: On community structure in
  complex networks: challenges and opportunities. Applied Network Science
  4(1),  1--35 (2019)

\bibitem{webkb}
Craven, M., DiPasquo, D., Freitag, D., McCallum, A., Mitchell, T., Nigam, K.,
  Slattery, S.: {Learning to extract symbolic knowledge from the World Wide
  Web}. Proceedings of the National Conference on Artificial Intelligence pp.
  509--516 (1998)

\bibitem{erdos1960evolution}
Erd{\H{o}}s, P., R{\'e}nyi, A.: On the evolution of random graphs. Publ. Math.
  Inst. Hung. Acad. Sci  5(1),  17--60 (1960)

\bibitem{graphsage}
Hamilton, W., Ying, Z., Leskovec, J.: Inductive representation learning on
  large graphs. In: Advances in Neural Information Processing Systems. pp.
  1024--1034 (2017)

\bibitem{hasani2018consensus}
Hasani-Mavriqi, I., Kowald, D., Helic, D., Lex, E.: Consensus dynamics in
  online collaboration systems. Computational social networks  5(1), ~2 (2018)

\bibitem{holland1983stochastic}
Holland, P.W., Laskey, K.B., Leinhardt, S.: Stochastic blockmodels: First
  steps. Social networks  5(2),  109--137 (1983)

\bibitem{karrer2011stochastic}
Karrer, B., Newman, M.E.: Stochastic blockmodels and community structure in
  networks. Physical review E  83(1),  016107 (2011)

\bibitem{kim2008complex}
Kim, J., Wilhelm, T.: What is a complex graph? Physica A: Statistical Mechanics
  and its Applications  387(11),  2637--2652 (2008)

\bibitem{gcn}
Kipf, T.N., Welling, M.: Semi-supervised classification with graph
  convolutional networks. In: International Conference on Learning
  Representations (ICLR) (2017)

\bibitem{Klicpera2019predict}
Klicpera, J., Bojchevski, A., G{\"{u}}nnemann, S.: {Predict then propagate:
  Graph neural networks meet personalized PageRank}. In: 7th International
  Conference on Learning Representations, ICLR 2019 (2019)

\bibitem{Li2018Deeper}
Li, Q., Han, Z., Wu, X.M.: {Deeper insights into graph convolutional networks
  for semi-supervised learning}. In: 32nd AAAI Conference on Artificial
  Intelligence, AAAI 2018 (2018)

\bibitem{Loukas2020What}
Loukas, A.: What graph neural networks cannot learn: depth vs width. In:
  International Conference on Learning Representations (2020),
  \url{https://openreview.net/forum?id=B1l2bp4YwS}

\bibitem{pubmed}
Namata, G., London, B., Getoor, L., Huang, B., EDU, U.: Query-driven active
  surveying for collective classification. In: 10th International Workshop on
  Mining and Learning with Graphs. vol.~8 (2012)

\bibitem{confmodel}
Newman, M.E.: The structure and function of complex networks. SIAM review
  45(2),  167--256 (2003)

\bibitem{press1992numerical}
Press, W.H., Teukolsky, S.A., Flannery, B.P., Vetterling, W.T.: Numerical
  recipes in Fortran 77: volume 1, volume 1 of Fortran numerical recipes: the
  art of scientific computing. Cambridge university press (1992)

\bibitem{twitter}
Ribeiro, M.H., Calais, P.H., Santos, Y.A., Almeida, V.A., Meira~Jr, W.: " like
  sheep among wolves": Characterizing hateful users on twitter. arXiv preprint
  arXiv:1801.00317  (2017)

\bibitem{sen2008collective}
Sen, P., Namata, G., Bilgic, M., Getoor, L., Galligher, B., Eliassi-Rad, T.:
  Collective classification in network data. AI magazine  29(3),  93--93 (2008)

\bibitem{shchur2018pitfalls}
Shchur, O., Mumme, M., Bojchevski, A., G{\"u}nnemann, S.: Pitfalls of graph
  neural network evaluation. Relational Representation Learning Workshop,
  NeurIPS 2018  (2018)

\bibitem{Tiao2019}
Tiao, L., Elinas, P., Nguyen, H., Bonilla, E.V.: {Variational Spectral Graph
  Convolutional Networks}. Graph Representation Learning Workshop, NeurIPS 2019
   (2019)

\bibitem{gat}
Veli{\v{c}}kovi{\'{c}}, P., Cucurull, G., Casanova, A., Romero, A., Li{\`{o}},
  P., Bengio, Y.: {Graph Attention Networks}. International Conference on
  Learning Representations  (2018),
  \url{https://openreview.net/forum?id=rJXMpikCZ}, accepted as poster

\bibitem{sgc}
Wu, F., Zhang, T., Souza~Jr, A.H.d., Fifty, C., Yu, T., Weinberger, K.Q.:
  Simplifying graph convolutional networks. arXiv preprint arXiv:1902.07153
  (2019)

\bibitem{Xu2019How}
Xu, K., Jegelka, S., Hu, W., Leskovec, J.: {How powerful are graph neural
  networks?} In: 7th International Conference on Learning Representations, ICLR
  2019 (2019)

\bibitem{zhu2003semi}
Zhu, X., Ghahramani, Z., Lafferty, J.D.: Semi-supervised learning using
  gaussian fields and harmonic functions. In: Proceedings of the 20th
  International conference on Machine learning (ICML-03). pp. 912--919 (2003)

\end{thebibliography}

\end{document}